\documentclass[10pt]{article} %
\usepackage[preprint]{tmlr}

\usepackage{amsmath,amsfonts,bm}

\def\eqref#1{equation~\ref{#1}}

\def\1{\bm{1}}

\DeclareMathAlphabet{\mathsfit}{\encodingdefault}{\sfdefault}{m}{sl}
\SetMathAlphabet{\mathsfit}{bold}{\encodingdefault}{\sfdefault}{bx}{n}

\usepackage{graphicx}
\usepackage{hyperref}
\usepackage{url}

\title{The Standard Deviation in VICReg Buys a Safety Margin}

\author{\name Hawar Dizayee \email hawardizayee@gmail.com \\
      \addr Independent Researcher}

\def\month{MM}  %
\def\year{YYYY} %
\def\openreview{\url{https://openreview.net/forum?id=XXXX}} %

\begin{document}

\maketitle

\begin{abstract}
VICReg prevents collapse with a hinge on the standard deviation of each embedding dimension, and its authors argue from the gradient that a hinge on the variance would fail. To our knowledge, this claim has not been tested experimentally. We test it in a controlled setting (ResNet-18, CIFAR-10 and CIFAR-100), changing only the hinge and sweeping its weight $\mu$ over five orders of magnitude. At the default weight ($\mu = 1$), the two hinges reach the same linear-probe accuracy on CIFAR-10 (about 89.5\%). Both collapse below a threshold weight, but the standard-deviation hinge's threshold is 35--130$\times$ lower. CIFAR-100 shows a consistent gap. In training, the variance hinge's gradient fades as the feature variance shrinks, while the standard-deviation hinge's does not until that variance is very small. The standard-deviation hinge therefore buys a safety margin: it keeps training at weights far below the point where the variance hinge collapses.
\end{abstract}

\section{Introduction}
\label{sec:intro}

VICReg~\citep{bardes2022vicreg} keeps a joint-embedding model from collapsing with three
loss terms: an invariance term that pulls the two views of an image together, a covariance
term that decorrelates embedding dimensions, and a variance term that keeps every dimension
spread out across the batch. The variance term is a hinge on the \emph{standard deviation}
of each dimension. \citet[Section~4.1]{bardes2022vicreg} state that using the standard
deviation rather than the variance is crucial: with a hinge on the variance, the gradient
vanishes as the embeddings shrink, and training collapses. This argument is made from the
form of the gradient; to our knowledge it has not been tested directly. The EB-JEPA
library~\citep{terver2026ebjepa}, which we build on, uses the same standard-deviation hinge.

We test the claim: we change only the hinge (standard deviation or variance) and sweep its
weight $\mu$ over five orders of magnitude. At EB-JEPA's default weight ($\mu = 1$) the two
hinges reach the same accuracy (about 89.5\% on CIFAR-10). Both collapse when the weight is
low enough, but the variance hinge collapses first, at a weight 35 to 130 times higher than
the standard-deviation hinge.

Our contributions are:
\begin{itemize}
  \item A controlled test of the standard-deviation-versus-variance claim in VICReg,
        changing only the hinge and sweeping its weight over five orders of magnitude,
        with three seeds at every weight and a CIFAR-100 check (Section~\ref{sec:results}).
  \item The finding that both hinges have a collapse threshold, that the two
        thresholds sit 35--130$\times$ apart, and that the hinges tie at the default
        weight (Section~\ref{sec:results}).
  \item An account of the mechanism from the measured gradients during training, including why the standard-deviation hinge also collapses
        (Section~\ref{sec:mechanism}).
  \item Code to reproduce every run, publicly available at \url{https://github.com/hawardizayee/vicreg-hinge}.
\end{itemize}

\section{Background}
\label{sec:background}
\paragraph{The VICReg loss.}
VICReg \citep{bardes2022vicreg} trains an encoder followed by a projector on two augmented
views of each image. Let $Z, Z' \in \mathbb{R}^{N \times D}$ be the projector outputs of a
batch of $N$ images for the two views, with rows $z_i$ and entries $z_{ij}$. We call each of the $D$ embedding dimensions a \emph{feature} and write $\bar{z}_j$ for the batch mean of feature $j$. The loss has three terms. The invariance term
pulls the two views of an image together:
\begin{equation}
s(Z, Z') = \frac{1}{ND} \sum_{i=1}^{N} \lVert z_i - z'_i \rVert^2 .
\label{eq:inv}
\end{equation}
The variance term keeps each feature from becoming constant over the batch. VICReg puts a
hinge on the standard deviation of each feature:
\begin{equation}
v_{\mathrm{std}}(Z) = \frac{1}{D} \sum_{j=1}^{D} \max\!\left(0,\; \gamma - \sqrt{\mathrm{Var}_j + \epsilon}\right),
\quad \text{where} \quad
\mathrm{Var}_j = \frac{1}{N-1} \sum_{i=1}^{N} (z_{ij} - \bar{z}_j)^2 .
\label{eq:std}
\end{equation}
The covariance term decorrelates the features:
\begin{equation}
c(Z) = \frac{1}{D(D-1)} \sum_{j \neq k} [C(Z)]_{j,k}^2 ,
\quad \text{where} \quad
C(Z) = \frac{1}{N-1} \sum_{i=1}^{N} (z_i - \bar{z})(z_i - \bar{z})^{\top} .
\label{eq:cov}
\end{equation}
The total loss is
\begin{equation}
\mathcal{L} = s(Z, Z') + \mu \left[ v(Z) + v(Z') \right] + \nu \left[ c(Z) + c(Z') \right],
\label{eq:loss}
\end{equation}
with $\gamma = 1$ and $\epsilon = 10^{-4}$ unless stated otherwise. We write the three terms as the EB-JEPA implementation computes them, with the invariance weight fixed to 1; they differ from the definitions in the VICReg paper by constant factors (Appendix~\ref{app:scaling}).

\paragraph{The variance hinge.}
The alternative we test puts the hinge on the variance itself:
\begin{equation}
v_{\mathrm{var}}(Z) = \frac{1}{D} \sum_{j=1}^{D} \max\!\left(0,\; \gamma - \mathrm{Var}_j\right).
\label{eq:var}
\end{equation}
With $\gamma = 1$ both hinges are zero for a feature of unit variance. For a constant
feature the variance hinge equals 1 and the standard-deviation hinge equals
$1 - \sqrt{\epsilon}$. They differ in their gradient (Section~\ref{sec:mechanism}, Eq.~\ref{eq:grads}).

\paragraph{The claim we test.}
\citet{bardes2022vicreg} state that using the standard deviation and not the variance in the
hinge is crucial (their Section 4.1). Their argument is about the gradient: with the
variance, the gradient with respect to a value goes to zero as the value approaches the
batch mean, so the term stops pushing the embeddings apart and they collapse.

\section{Setup}
\label{sec:setup}

\paragraph{Code and model.}
All runs use a standalone reimplementation of the \texttt{image\_jepa} example of the EB-JEPA library~\citep{terver2026ebjepa,ebjepa}, with three flags added: \texttt{loss.hinge=std|var}, which picks the variance hinge (Section~\ref{sec:background}), \texttt{loss.eps}, which sets the $\epsilon$ of the standard-deviation hinge, and a dataset flag for CIFAR-100. The encoder is a ResNet-18~\citep{he2016resnet} with the usual CIFAR stem, followed by a projector of three linear layers (hidden and output width 1024), with batch normalization and ReLU after the first two. The loss is applied to the projector outputs. The code is publicly available at \url{https://github.com/hawardizayee/vicreg-hinge}.

\paragraph{Data and training.}
We train on CIFAR-10~\citep{krizhevsky2009cifar} for 300 epochs with batch size 256, using LARS~\citep{you2017lars} (learning rate 0.3, momentum 0.9, weight decay $10^{-4}$, $\eta = 0.02$) with 10 warmup epochs and cosine decay to 0, in bfloat16 mixed precision. Augmentations are the EB-JEPA defaults (random resized crop, color jitter, grayscale, solarization, horizontal flip), applied independently to each view. We repeat part of the grid on CIFAR-100, normalized with CIFAR-100's own mean and standard deviation. Full details are in Appendix~\ref{app:impl}.

\paragraph{Loss weights.}
The invariance term has weight 1, as in EB-JEPA, and the covariance weight is 100. Both are fixed in every run of the main grid. The only loss weight we change is the variance weight $\mu$. These weights cannot be compared directly to the 25\,/\,25\,/\,1 of the VICReg paper~\citep{bardes2022vicreg}; Appendix~\ref{app:scaling} converts them.

\paragraph{What we vary.}
Runs in the main grid differ in exactly three things: the hinge (std or var), $\mu$, and the seed.
Tables~\ref{tab:main} and~\ref{tab:c100} list every weight of the grid for each hinge; the weights are
denser near each collapse threshold, and the grid includes runs with no variance term ($\mu = 0$).
Every point, on CIFAR-10 and on CIFAR-100, uses three seeds (42, 1, 1000).
Three controls step outside this grid and are described where they are reported: full fp32,
VICReg's covariance weight, and a larger $\epsilon$.

\paragraph{Evaluation.}
We report the accuracy of an online linear probe trained on the encoder features during pre-training, following the EB-JEPA protocol. The probe trains on detached encoder features. The number we report is validation accuracy after the last epoch. A run is \emph{collapsed} when this accuracy ends near the no-variance-term baseline (16--23\% on CIFAR-10, against 10\% chance), and \emph{diverged} when the loss becomes NaN. Collapsed runs stay above the 10\% chance level because collapse is not exact: the encoder outputs of different images still differ slightly, and the probe can use those differences.

\paragraph{Training-dynamics monitor.}
For Section~\ref{sec:mechanism}, we fix a monitor batch: the first 256 training images, each with two augmented views drawn once before training. After every epoch, we pass both views through the network in evaluation mode, in fp32 and without tracking gradients, and save the encoder and projector outputs. The feature-scale and gradient analyses are both computed afterward from these saved outputs.

\paragraph{Compute.}
Each run takes about 65 minutes on one NVIDIA RTX PRO 4500.
\section{Results}
\label{sec:results}

\begin{figure}[t]
\centering
\includegraphics[width=\linewidth]{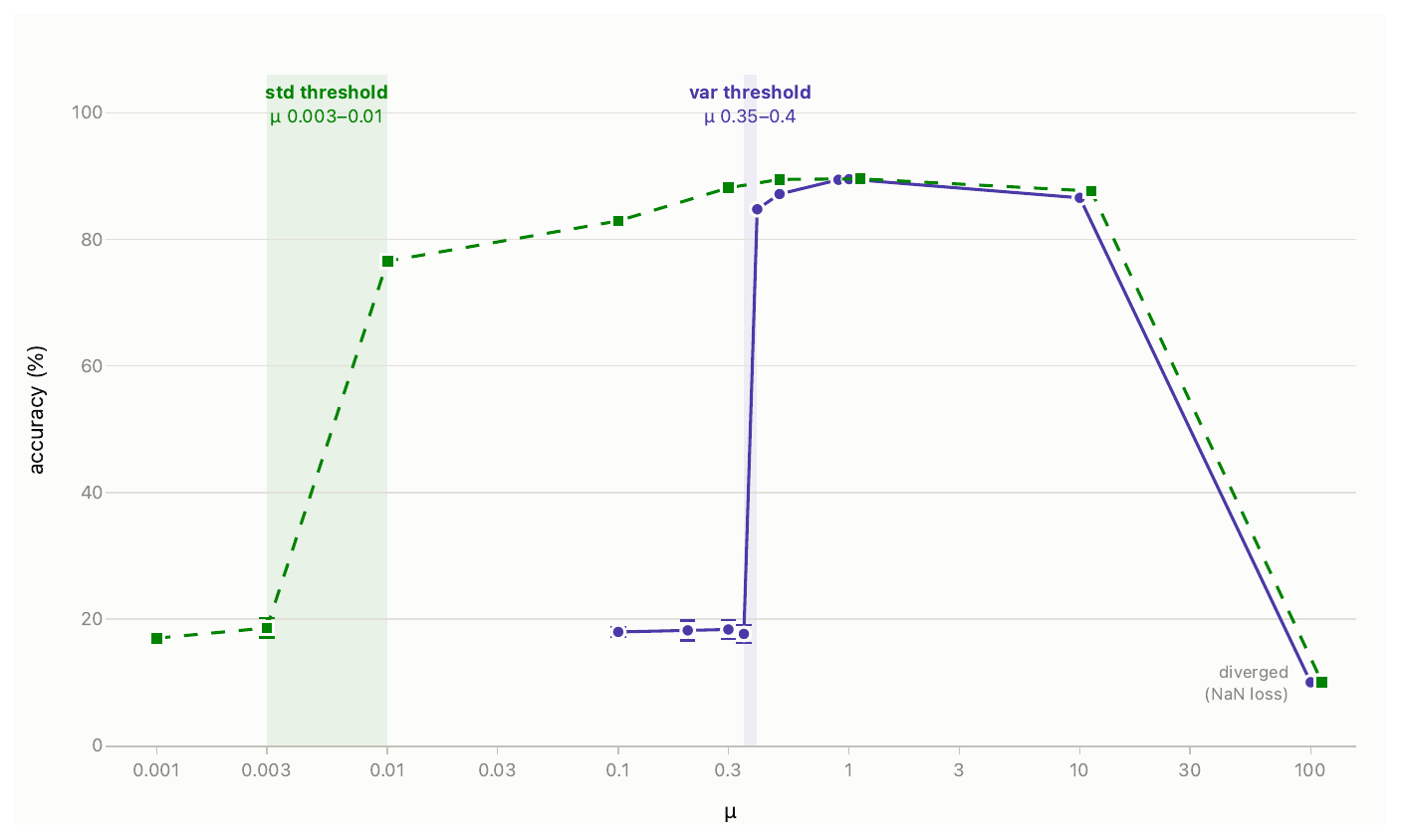}
\caption{CIFAR-10 linear-probe accuracy after 300 epochs versus the variance-term weight
$\mu$, for the variance hinge (var; violet circles, solid line) and the standard-deviation
hinge (std; green squares, dashed line). Each point is the mean over 3 seeds; error bars
show the standard deviation and are often smaller than the marker. Shaded bands mark each
hinge's collapse threshold (std: 0.003--0.01; var: 0.35--0.4). At $\mu = 100$ training
diverged for both hinges (NaN loss; the probe is at chance, 10\%). Where the two hinges
nearly coincide ($\mu = 1$, $10$ and $100$) the std marker is shifted slightly to the right
for visibility.}
\label{fig:main}
\end{figure}

\begin{table}[t]
\small
\caption{Linear-probe accuracy (\%) on CIFAR-10 at epoch 300 for the variance (var) and
standard-deviation (std) hinge across the variance-term weight $\mu$.
Mean over 3 seeds (1, 42, 1000), with the standard deviation in small type.
Collapsed runs end between 16\% and 23\% (chance 10\%), the level of the run with no
variance term; Table~\ref{tab:full} gives their values.
$^\ddagger$ training diverged (loss became NaN; probe at chance).
-- : not run.}
\label{tab:main}
\begin{center}
\begin{tabular}{lcc}
\hline
$\mu$ & var hinge & std hinge \\
\hline
0 (no var term) & \multicolumn{2}{c}{18.79\,{\scriptsize$\pm$\,3.06}} \\
0.001 & -- & collapsed \\
0.003 & -- & collapsed \\
0.01 & -- & 76.57\,{\scriptsize$\pm$\,0.21} \\
0.1 & collapsed & 82.93\,{\scriptsize$\pm$\,0.61} \\
0.2 & collapsed & -- \\
0.3 & collapsed & 88.15\,{\scriptsize$\pm$\,0.17} \\
0.35 & collapsed & -- \\
0.4 & 84.77\,{\scriptsize$\pm$\,0.33} & -- \\
0.5 & 87.17\,{\scriptsize$\pm$\,0.16} & 89.48\,{\scriptsize$\pm$\,0.25} \\
0.9 & 89.43\,{\scriptsize$\pm$\,0.09} & -- \\
1 & 89.53\,{\scriptsize$\pm$\,0.13} & 89.58\,{\scriptsize$\pm$\,0.15} \\
10 & 86.58\,{\scriptsize$\pm$\,0.05} & 87.65\,{\scriptsize$\pm$\,0.17} \\
100 & div.$^\ddagger$ & div.$^\ddagger$ \\
\hline
\end{tabular}
\end{center}
\end{table}

Figure~\ref{fig:main} and Table~\ref{tab:main} give the linear-probe accuracy at the end of
training for every weight $\mu$ and both hinges. At no weight
do the three seeds disagree on the outcome.

\paragraph{The hinges tie at the default weight.}
At $\mu = 1$ the var hinge reaches $89.53 \pm 0.13$ and the std hinge $89.58 \pm 0.15$ on
CIFAR-10. The gap of 0.05 points is smaller than the spread over seeds. On CIFAR-100
the std hinge is half a point ahead ($63.82 \pm 0.27$ against $63.33 \pm 0.24$;
Table~\ref{tab:c100}), about twice the spread over seeds. A hinge on the
variance is therefore not a failing design in itself: at the default weight it trains to within half a point of the hinge on the standard deviation on both datasets.

\paragraph{Both hinges have a collapse threshold, far apart.}
The var threshold lies
between 0.35 and 0.4 and the std threshold between 0.003 and 0.01, which is 35 to 130 times
lower.

\begin{figure}[t]
\centering
\includegraphics[width=\linewidth]{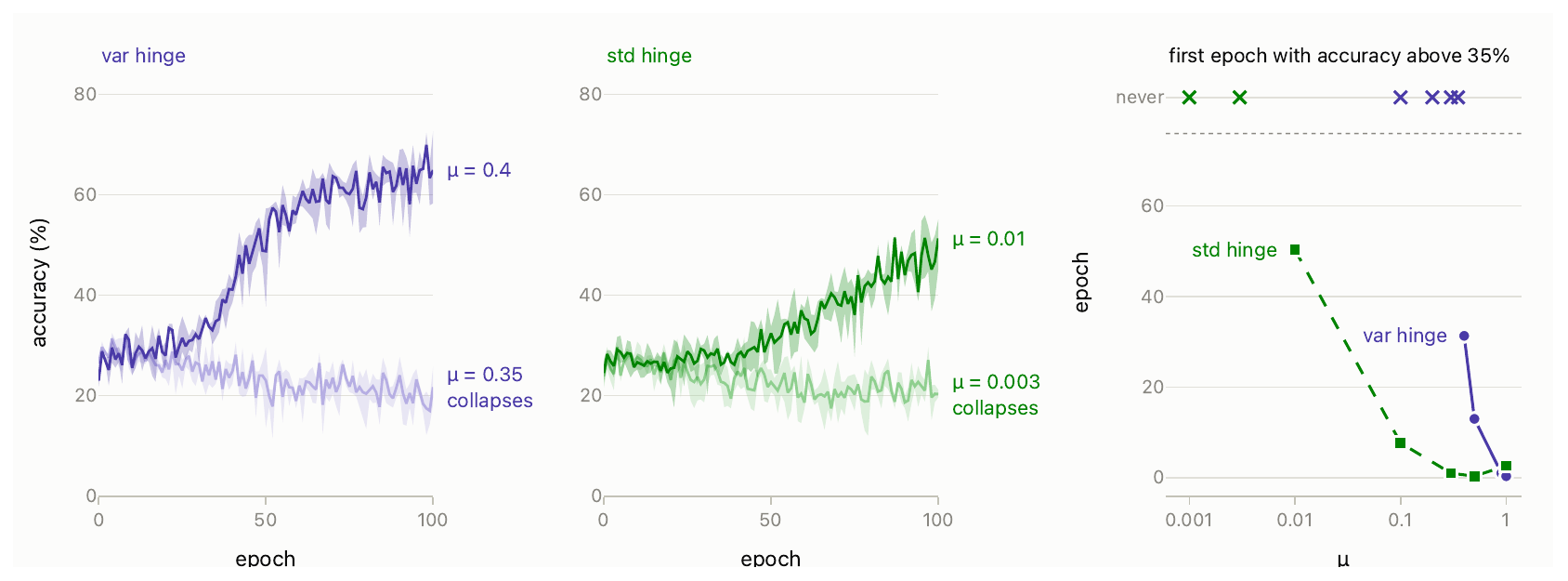}
\caption{The collapse threshold over the first 100 epochs of training (CIFAR-10). Colors
as in Figure~\ref{fig:main}. \textbf{Left and middle:} linear-probe accuracy per epoch on
the two sides of each hinge's threshold. Lines are the mean over 3 seeds and the shaded band
spans the lowest to the highest seed; the paler curve is the weight that collapses.
\textbf{Right:} escape delay, the first epoch at which accuracy exceeds 35\%, against $\mu$
(mean over 3 seeds). A cross marks a weight at which all three seeds collapse and never
reach 35\%.}
\label{fig:cliff}
\end{figure}

\paragraph{Close to a threshold: lower accuracy and a late start.}
Lowering $\mu$ costs accuracy for both hinges before it causes collapse, the var hinge sooner
than the std hinge (2.3 points behind at $\mu = 0.5$; Tables~\ref{tab:main}
and~\ref{tab:c100}), and the effective rank of the final representations falls with it
(Appendix~\ref{app:rankme}). Training also starts later.
Figure~\ref{fig:cliff} (left and middle) shows the two sides of each threshold over the
first 100 epochs: just above a threshold, the surviving runs cannot be told apart from the
collapsing ones until about epoch 35 for the var hinge ($\mu = 0.4$) and about epoch 50 for the std
hinge ($\mu = 0.01$). We call the first epoch at which a run's accuracy exceeds 35\%, a level
no collapsed run reaches, its \emph{escape delay}. It grows as $\mu$ approaches a threshold
from above, for both hinges (Figure~\ref{fig:cliff}, right), so a run at a low weight can look
collapsed for tens of epochs and still train.

\begin{table}[t]
\small
\caption{Linear-probe accuracy (\%) on CIFAR-100 at epoch 300, in the format of
Table~\ref{tab:main}.
Collapsed runs end between 2\% and 4\% (chance 1\%); Table~\ref{tab:full} gives
their values. -- : not run.}
\label{tab:c100}
\begin{center}
\begin{tabular}{lcc}
\hline
$\mu$ & var hinge & std hinge \\
\hline
0.003 & -- & collapsed \\
0.01 & -- & 35.74\,{\scriptsize$\pm$\,0.29} \\
0.1 & collapsed & 54.52\,{\scriptsize$\pm$\,0.54} \\
0.3 & collapsed & -- \\
0.4 & 56.34\,{\scriptsize$\pm$\,0.55} & -- \\
0.5 & 58.90\,{\scriptsize$\pm$\,0.25} & -- \\
1 & 63.33\,{\scriptsize$\pm$\,0.24} & 63.82\,{\scriptsize$\pm$\,0.27} \\
\hline
\end{tabular}
\end{center}
\end{table}

\paragraph{CIFAR-100.}
On CIFAR-100 both thresholds are consistent with those on CIFAR-10 (chance 1\%; Table~\ref{tab:c100}). The
var hinge collapses at $\mu = 0.1$ ($3.26 \pm 0.35$) and $\mu = 0.3$
($3.23 \pm 0.29$) and survives at $\mu = 0.4$ ($56.34 \pm 0.55$), so its threshold lies between
0.3 and 0.4. The std hinge collapses at $\mu = 0.003$ ($3.37 \pm 0.42$) and
survives at $\mu = 0.01$ ($35.74 \pm 0.29$), so its threshold lies between 0.003 and 0.01. Just above its threshold the std hinge is far below its
accuracy at $\mu = 0.1$ ($54.52 \pm 0.54$), a larger loss than on CIFAR-10.

\paragraph{Controls.}
Training uses mixed precision, and the hinge computes a batch variance at every step, so we
checked that neither threshold is a precision effect. Rerun in full fp32 (seed 42), the
var hinge still collapses at $\mu = 0.3$ (15.62) and $\mu = 0.35$ (15.79), and the std hinge
still collapses at $\mu = 0.003$ (15.83) and survives at $\mu = 0.01$ (76.66).
We also reran the std hinge with $\epsilon = 10^{-2}$ (seed 42): it collapses at $\mu = 0.05$
and survives at $\mu = 0.1$, so its threshold moves up roughly tenfold, as
Section~\ref{sec:mechanism} predicts.
At VICReg's default ratios ($\mu = 0.5$, $\nu = 41$ in our units) both hinges train and the
std hinge is 0.57 points ahead (Appendix~\ref{app:scaling}). Above $\mu = 1$ both hinges lose accuracy at $\mu = 10$, the var hinge about a point more, and both diverge at $\mu = 100$ (consistent with Table~2 of \citealt{terver2026ebjepa}). This high-weight regime is a different failure from low-$\mu$ collapse and is outside this study.

\section{Mechanism}
\label{sec:mechanism}

\begin{figure}[t]
\centering
\includegraphics[width=\linewidth]{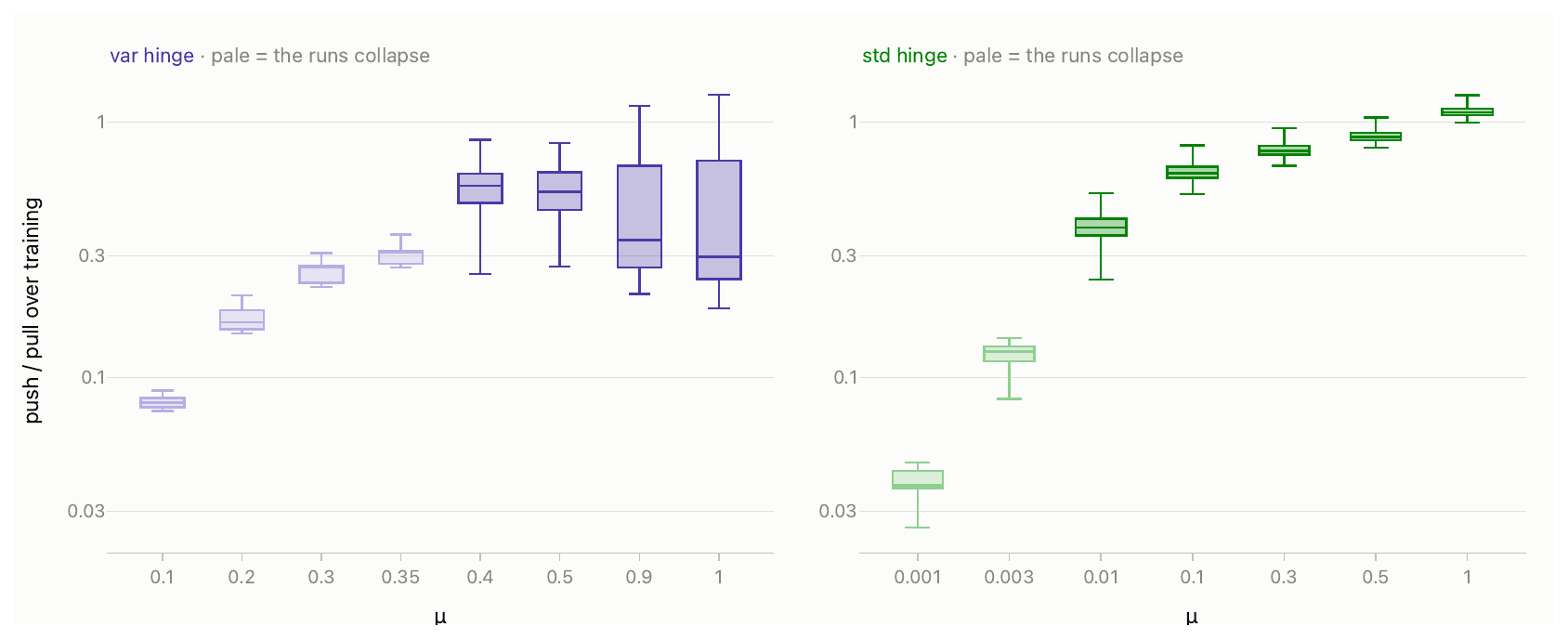}
\caption{Push/pull over training (CIFAR-10), on a logarithmic axis: the norm of the gradient
of the weighted variance term divided by the norm of the gradient of the invariance term,
both taken with respect to the projector outputs of the monitor batch. One box per weight
$\mu$, holding every epoch of the 3 seeds (900 values). The box spans the middle half of the
values, the line is the median and the whiskers reach the lowest and the highest value. Pale
boxes are weights that collapse; colors as in Figure~\ref{fig:main}.}
\label{fig:ratio}
\end{figure}

\paragraph{Push and pull.}
The invariance term pulls the two views of an image together, and the variance term pushes
each feature away from its batch mean. For a feature $j$ below the target $\gamma$, the
gradients of the two hinges, before the weight $\mu$, are
\begin{equation}
\frac{\partial v_{\mathrm{var}}}{\partial z_{ij}} = -\frac{2}{D(N-1)} \,(z_{ij} - \bar{z}_j),
\qquad
\frac{\partial v_{\mathrm{std}}}{\partial z_{ij}} = -\frac{1}{D(N-1)} \,\frac{z_{ij} - \bar{z}_j}{\sqrt{\mathrm{Var}_j + \epsilon}} ,
\label{eq:grads}
\end{equation}
and the gradient of the invariance term is
$\partial s / \partial z_{ij} = \frac{2}{ND}(z_{ij} - z'_{ij})$. At every epoch we take the
gradient of the weighted variance term and the gradient of the invariance term with respect to
the projector outputs of the monitor batch, and divide the norm of the first by the norm of the second, each taken over all entries of both views. We call this ratio push/pull (Figure~\ref{fig:ratio}).

\paragraph{The two hinges push differently.}
The var hinge pushes in proportion to the deviation $z_{ij}-\bar z_j$, so its push shrinks with
the feature. This is the argument of VICReg. The pull shrinks with the features as well, so for
the var hinge push/pull hardly changes as a run collapses: it is set by $\mu$. The std hinge divides the deviation by
$S_j = \sqrt{\mathrm{Var}_j + \epsilon}$, so its push keeps its size as the feature shrinks and
push/pull rises, until $S_j$ reaches its floor $\sqrt{\epsilon} = 0.01$. For both hinges the ratio falls as $\mu$ decreases, except for the var hinge at $\mu \ge 0.9$, discussed below (Figure~\ref{fig:ratio}).

\paragraph{Why the std hinge's threshold is lower.}
Because its push does not shrink, the std hinge gains on the pull as a feature shrinks, and
even a weak hinge can win. This holds until the spread falls below $\sqrt{\epsilon}$. There
$S_j \approx \sqrt{\epsilon}$, and the std hinge of Eq.~\ref{eq:grads} becomes a var hinge with
a push $1/(2\sqrt{\epsilon}) = 50$ times larger. The std
hinge's threshold therefore depends on $\epsilon$. A one-seed control with $\epsilon = 10^{-2}$ agrees: the factor falls to 5 and the threshold moves up about tenfold (Section~\ref{sec:results}).

\paragraph{A low ratio late in training is not collapse.}
The tall var boxes at $\mu \ge 0.9$ have a different cause. There most features reach the
target variance of 1 during training. The hinge is zero for those features, so the push
falls: at $\mu = 1$ (seed 42) only 15\% of the features are still below the target at the end
of training. Push/pull describes the tug-of-war only while the hinge is active for nearly all
features, which holds for every collapsing run.

\section{Related Work}
\label{sec:related}
Joint-embedding methods must keep their embeddings from collapsing to a constant, and they do so in different ways. BYOL~\citep{grill2020byol} and SimSiam~\citep{chen2021simsiam} rely on an asymmetric architecture with a predictor and a stop-gradient. Barlow Twins~\citep{zbontar2021barlow} drives the cross-correlation matrix of the two views toward the identity, and VICReg~\citep{bardes2022vicreg} regularizes the variance and the covariance of each view directly. Collapse can also be partial: \citet{jing2022dimensional} show that the embeddings can end up spanning a lower-dimensional subspace, and RankMe~\citep{garrido2023rankme}, which we report in Appendix~\ref{app:rankme}, measures the effective rank of the embeddings. Analyses of VICReg relate it to contrastive methods~\citep{garrido2023duality} and to the optimization of mutual information~\citep{shwartzziv2023information}.

\citet{bardes2022vicreg} argue for the standard deviation from the form of the gradient, and
their ablations vary the loss weights, not the hinge. The EB-JEPA
paper~\citep{terver2026ebjepa} sweeps the variance weight on CIFAR-10, again with the
standard-deviation hinge and only at weights of 1 and above, and notes that VICReg needs more
careful tuning than SIGReg~\citep{balestriero2025lejepa}. We found no work that trains with a hinge on the variance, or that brackets the weight at which either hinge collapses. The closest theoretical result is that
VICReg does not in general recover full-rank representations~\citep{chapman2024cca}; it
concerns the objective, not the hinge.

LeJEPA~\citep{balestriero2025lejepa} replaces the variance and covariance terms with SIGReg,
which has a single trade-off weight. VISReg~\citep{wu2026visreg} keeps a standard-deviation
term for scale, with the same gradient argument as VICReg. We tested neither.

\section{Limitations}
The thresholds we report hold for one setup (ResNet-18, CIFAR, $\epsilon = 10^{-4}$, fixed
invariance and covariance weights). The standard-deviation hinge's threshold depends on
$\epsilon$, which we varied once and on one seed.
Both thresholds are also specific to the loss scaling of EB-JEPA (Appendix~\ref{app:scaling}). Our claim
is the pattern, not the values of $\mu$ at which it occurs. All accuracies come from an online linear probe at a fixed budget of 300 epochs; we did not run an offline linear evaluation or a k-nearest-neighbor evaluation.

Push/pull compares the sizes of two gradients on one batch, not their directions or the
resulting weight updates, so it is a diagnostic consistent with the gradient argument, not a
proof of it.

\section{Conclusion}
VICReg's authors argue that the hinge must act on the standard deviation, because a hinge on the variance loses its gradient as the embeddings shrink. In our setting, the experiments support the gradient argument but narrow what follows from it. A hinge on the variance trains at the default weight, and both hinges collapse below a threshold weight, with the standard-deviation hinge's threshold 35 to 130 times lower. What the standard deviation buys is a safety margin: a larger distance between the default weight and the collapse threshold. The measured gradients are consistent with this: the variance hinge's push fades as the features shrink, while the standard-deviation hinge's push holds until the spread falls below $\sqrt{\epsilon}$.

\bibliography{main}
\bibliographystyle{tmlr}

\clearpage
\appendix
\section{Relation to the weights of the VICReg paper}
\label{app:scaling}

The weights in Eq.~\ref{eq:loss} are those of the EB-JEPA implementation. They are not on the
scale of the $(\lambda, \mu, \nu)$ of \citet{bardes2022vicreg}, because the terms are
normalized differently. Table~\ref{tab:terms} gives each term as written in the VICReg paper,
as computed by the official VICReg code~\citep{vicregcode}, and as computed by EB-JEPA~\citep{ebjepa}.

\begin{table}[ht]
\small
\caption{The three loss terms as written in the VICReg paper, as computed by the official
VICReg code, and as computed by EB-JEPA. $D$ is the embedding size.}
\label{tab:terms}
\begin{center}
\begin{tabular}{llll}
\hline
Term & VICReg paper & Official VICReg code & EB-JEPA \\
\hline
Invariance & summed over dimensions, & averaged over both & as the code \\
 & averaged over images & & \\
Variance & the two views added & the two views averaged & the two views added \\
Covariance & squared off-diagonals & as the paper & averaged over the \\
 & summed, divided by $D$ & & $D(D-1)$ off-diagonals \\
\hline
\end{tabular}
\end{center}
\end{table}

The paper and its code differ in the first two rows. The weights $(25, 25, 1)$ are the
defaults of the code, so we convert against the code.

Dividing the code's loss by $\lambda$ and comparing it with Eq.~\ref{eq:loss} term by term
gives $\mu_{\mathrm{ours}} = \mu / (2\lambda)$ and
$\nu_{\mathrm{ours}} = \nu\,(D - 1) / \lambda$, with $D = 1024$ here. We convert only the
ratios to the invariance weight; the overall scale of the loss, which also sets its balance
against weight decay, is left as it is.

\begin{table}[ht]
\small
\caption{Weights of the VICReg paper and the same weights in our units, for $D = 1024$.}
\label{tab:conversion}
\begin{center}
\begin{tabular}{ll}
\hline
VICReg $(\lambda, \mu, \nu)$ & Ours $(\mu, \nu)$ \\
\hline
$(25, 25, 1)$, the default & $(0.5, 41)$ \\
$(10, 10, 1)$ & $(0.5, 102)$ \\
about $(10, 20, 1)$, not in their table & $(1, 100)$, our default \\
\hline
\end{tabular}
\end{center}
\end{table}

In every row of their Table~7 (arXiv version) that trains, the variance weight equals the invariance weight,
which is $\mu = 0.5$ in our units. EB-JEPA's default, $\mu = 1$, is twice that. We ran both
hinges at VICReg's default ratios, $\mu = 0.5$ and $\nu = 41$, with three seeds. Both train:
the std hinge reaches $89.16 \pm 0.13$ and the var hinge $88.59 \pm 0.19$, with the std hinge
ahead on every seed. At $\mu = 0.5$ and $\nu = 100$, which has to within 2\% the ratios of
their row $(10, 10, 1)$, the gap is larger: $89.48 \pm 0.25$ against $87.17 \pm 0.16$
(Table~\ref{tab:main}). We did not measure the var threshold at $\nu = 41$.

The conversion matches the loss only. VICReg's experiments use ImageNet, a ResNet-50, an
8192-wide projector and a batch size of 2048, so equal weights do not make the two settings
equivalent, and the values of $\mu$ we report should not be read as thresholds for VICReg's
own setup.

\section{Implementation details}
\label{app:impl}

Table~\ref{tab:impl} lists the settings shared by all runs of the main grid. They are those of the \texttt{image\_jepa} example of EB-JEPA~\citep{ebjepa}, which our code reimplements, except for the projector width (1024 here, 2048 by default in EB-JEPA). For this
configuration (batch size 256, 1024$\times$1024 projector, standard-deviation hinge, $\mu = 1$,
$\nu = 100$) the EB-JEPA repository reports 90.12\%; our reimplementation reaches
$89.58 \pm 0.15$ over three seeds (Table~\ref{tab:main}).

\begin{table}[ht]
\small
\caption{Settings shared by all runs of the main grid. The loss weights are given in Section~\ref{sec:setup}.}
\label{tab:impl}
\begin{center}
\begin{tabular}{lp{0.72\linewidth}}
\hline
Encoder & ResNet-18; first convolution $3 \times 3$ with stride 1 and padding 2, no max-pooling, no classification layer; output size 512 \\
Projector & three linear layers with hidden width 1024 and output width 1024; batch normalization and ReLU after the first two \\
Loss & computed on the projector outputs of the two views, under mixed precision \\
Optimizer & LARS, momentum 0.9, weight decay $10^{-4}$, $\eta = 0.02$ with learning-rate clipping; biases and normalization parameters are excluded from the LARS scaling and from weight decay \\
Learning rate & 0.3, set once per epoch: linear warmup from $3 \times 10^{-5}$ over 10 epochs, then cosine decay to 0 \\
Batch size, epochs & 256 (last incomplete batch dropped), 300 \\
Precision & bfloat16 automatic mixed precision \\
Augmentations & per view: random resized crop to $32 \times 32$ (scale 0.2--1), color jitter (brightness 0.4, contrast 0.4, saturation 0.2, hue 0.1; probability 0.8), grayscale (0.2), solarization (0.1), horizontal flip (0.5), then normalization with the dataset mean and standard deviation \\
Linear probe & one linear layer on the detached encoder output of the first view, cross-entropy loss, trained by the same optimizer and schedule as the model; evaluated on the 10,000 validation images (CIFAR's test split) without augmentation \\
Software & PyTorch 2.8.0 \\
\hline
\end{tabular}
\end{center}
\end{table}

\section{Accuracy at every weight}
\label{app:full}

Table~\ref{tab:full} gives the accuracy at every weight $\mu$, including the values of the
collapsed runs, which Tables~\ref{tab:main} and~\ref{tab:c100} only mark as collapsed.

\begin{table}[ht]
\small
\caption{Linear-probe accuracy (\%) at epoch 300 for every weight $\mu$, with the values of the
runs that Tables~\ref{tab:main} and~\ref{tab:c100} mark as collapsed.
Mean over 3 seeds (1, 42, 1000), with the standard deviation in small type.
\textit{Italic}: collapsed run (CIFAR-10 chance 10\%, CIFAR-100 chance 1\%).
$^\ddagger$ training diverged (loss became NaN; probe at chance).
-- : not run.}
\label{tab:full}
\begin{center}
\begin{tabular}{lcccc}
\hline
 & \multicolumn{2}{c}{CIFAR-10} & \multicolumn{2}{c}{CIFAR-100} \\
$\mu$ & var hinge & std hinge & var hinge & std hinge \\
\hline
0 (no var term) & \multicolumn{2}{c}{\textit{18.79\,{\scriptsize$\pm$\,3.06}}} & -- & -- \\
0.001 & -- & \textit{16.97\,{\scriptsize$\pm$\,0.29}} & -- & -- \\
0.003 & -- & \textit{18.58\,{\scriptsize$\pm$\,1.52}} & -- & \textit{3.37\,{\scriptsize$\pm$\,0.42}} \\
0.01 & -- & 76.57\,{\scriptsize$\pm$\,0.21} & -- & 35.74\,{\scriptsize$\pm$\,0.29} \\
0.1 & \textit{17.95\,{\scriptsize$\pm$\,0.76}} & 82.93\,{\scriptsize$\pm$\,0.61} & \textit{3.26\,{\scriptsize$\pm$\,0.35}} & 54.52\,{\scriptsize$\pm$\,0.54} \\
0.2 & \textit{18.20\,{\scriptsize$\pm$\,1.57}} & -- & -- & -- \\
0.3 & \textit{18.34\,{\scriptsize$\pm$\,1.47}} & 88.15\,{\scriptsize$\pm$\,0.17} & \textit{3.23\,{\scriptsize$\pm$\,0.29}} & -- \\
0.35 & \textit{17.64\,{\scriptsize$\pm$\,1.45}} & -- & -- & -- \\
0.4 & 84.77\,{\scriptsize$\pm$\,0.33} & -- & 56.34\,{\scriptsize$\pm$\,0.55} & -- \\
0.5 & 87.17\,{\scriptsize$\pm$\,0.16} & 89.48\,{\scriptsize$\pm$\,0.25} & 58.90\,{\scriptsize$\pm$\,0.25} & -- \\
0.9 & 89.43\,{\scriptsize$\pm$\,0.09} & -- & -- & -- \\
1 & 89.53\,{\scriptsize$\pm$\,0.13} & 89.58\,{\scriptsize$\pm$\,0.15} & 63.33\,{\scriptsize$\pm$\,0.24} & 63.82\,{\scriptsize$\pm$\,0.27} \\
10 & 86.58\,{\scriptsize$\pm$\,0.05} & 87.65\,{\scriptsize$\pm$\,0.17} & -- & -- \\
100 & div.$^\ddagger$ & div.$^\ddagger$ & -- & -- \\
\hline
\end{tabular}
\end{center}
\end{table}

\clearpage
\section{Effective rank of the final representations}
\label{app:rankme}

\begin{figure}[ht]
\centering
\includegraphics[width=\linewidth]{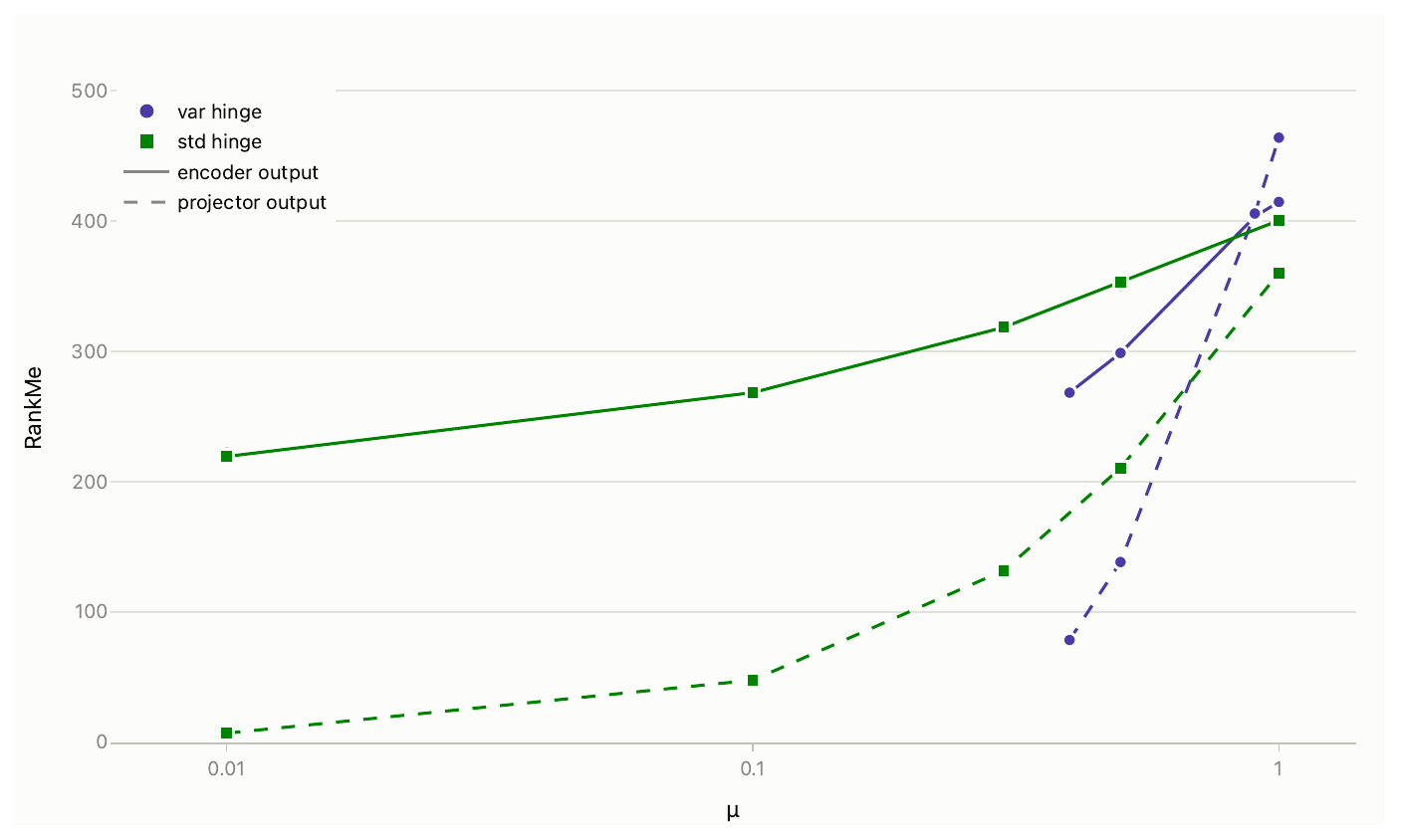}
\caption{RankMe \citep{garrido2023rankme} of the final model against the weight $\mu$
(CIFAR-10): the exponential of the entropy of the normalized singular values of the
embeddings of the 10,000 validation images, after subtracting their mean. Solid lines are
the encoder output (512 dimensions, so at most 512) and dashed lines the projector output
(1024 dimensions, at most 1024). Each point is the mean of the 3 seeds, which lie within 7 of
each other at every point; colors as in Figure~\ref{fig:main}. Weights at which the runs
collapse (var hinge $\mu \le 0.35$, std hinge $\mu \le 0.003$) are left out: their embeddings
are at numerical resolution and have no meaningful rank.}
\label{fig:rankme}
\end{figure}

Figure~\ref{fig:rankme} shows the effective rank of the final representations. For both hinges it falls as $\mu$ approaches the collapse threshold from above, and it falls further in the projector output than in the encoder output.

\end{document}